\documentclass[11pt,a4paper]{article}
\usepackage{acl2019}
\usepackage{times}
\usepackage{latexsym}
\usepackage{graphicx}
\usepackage{subcaption}
\usepackage{amssymb}
\usepackage{mathtools}
\usepackage{amsmath}
\usepackage{comment}
\usepackage{url}
\usepackage[para]{footmisc}

\aclfinalcopy 
\def\aclpaperid{1371} 

\title{Learning Attention-based Embeddings for Relation Prediction in Knowledge Graphs}

\newcommand*{\affaddr}[1]{#1} 
\newcommand*{\email}[1]{\texttt{#1}}

\author{%
  Deepak Nathani\thanks{\hspace{0.01in} Equal Contribution} \qquad
  Jatin Chauhan\footnotemark[1]  \qquad
  Charu Sharma\footnotemark[1]  \qquad
  Manohar Kaul \\
  \affaddr{Department of Computer Science and Engineering, IIT Hyderbad} \\
 \email{\{deepakn1019,chauhanjatin100,charusharma1991\}@gmail.com, mkaul@iith.ac.in}
  }

\date{}

\begin{document}
\maketitle
\begin{abstract}
The recent proliferation of knowledge graphs (KGs) coupled with incomplete or partial information, in the form of missing relations (links) between entities, has fueled a lot of research on knowledge base completion (also known as relation prediction). Several recent works suggest that convolutional neural network (CNN) based models generate richer and more expressive feature embeddings and hence also perform well on relation prediction. 
However, we observe that these KG embeddings treat triples independently and thus fail to cover the complex and hidden information that is inherently implicit in the local neighborhood surrounding a triple. 
To this effect, our paper proposes a novel attention-based feature embedding that captures both entity and relation features in any given entity's neighborhood. Additionally, we also encapsulate relation clusters and multi-hop relations 
in our model. Our empirical study offers insights into the efficacy of our attention-based model and we show marked performance gains in comparison to state-of-the-art methods on all datasets.

\end{abstract}

\section{Introduction}
\label{sec:intro}
\begin{figure}
	\includegraphics[width=\linewidth]{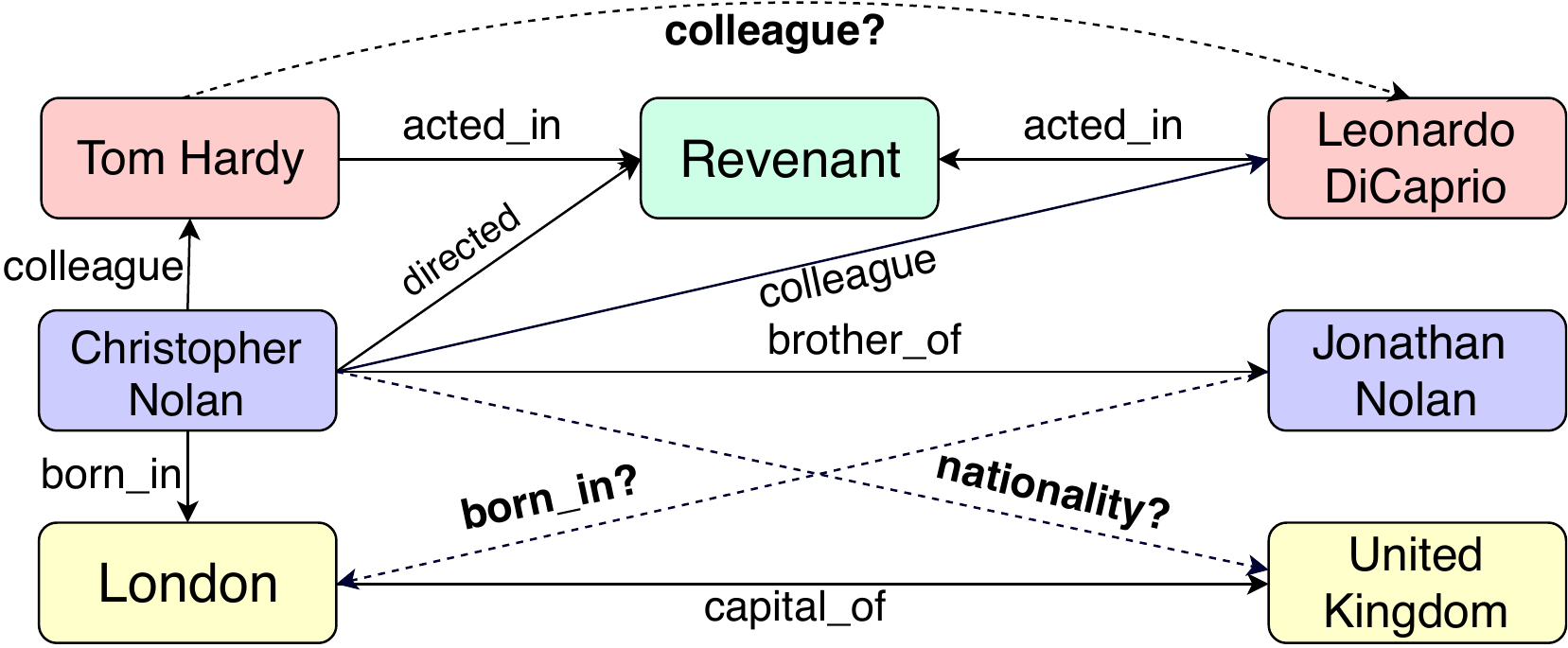}
	\caption{Subgraph of a knowledge graph contains actual relations between entities (solid lines) and inferred relations that are initially hidden (dashed lines).}
	\label{fig:toys}
\end{figure}

Knowledge graphs (KGs) represent knowledge bases (KBs) as a directed graph 
whose \emph{nodes} and \emph{edges} represent \emph{entities} and \emph{relations between entities}, respectively.
For example, in Figure ~\ref{fig:toys}, a triple \textit{(London, capital\_of, United Kingdom)} is represented as two entities: \textit{London} and \textit{United Kingdom} along with a relation \textit{(capital\_of)} linking them. 
KGs find uses in a wide variety of applications such as semantic search~\cite{berant2013semantic, berant2014}, dialogue generation~\cite{Hehe2017, keizer2017}, and question answering~\cite{zhang2016question,diefenbach2018wdaqua}, to name a few. However, KGs typically suffer from missing relations~\cite{Socher2013,West_2014}. 
This problem gives rise to the task of \emph{knowledge base completion} (also referred to as \emph{relation prediction}), which entails predicting whether a given triple is valid or not.

State-of-the-art relation prediction methods are known to be primarily \emph{knowledge embedding} based models. They are broadly classified as \emph{translational} models~\cite{NIPS2013_5071,yang2014,trouillon2016complex} and \emph{convolutional neural network} (CNN)~\cite{nguyen2018novel,dettmers2018convolutional} based models. 
While translational models learn embeddings using simple operations  and limited parameters, 
they produce low quality embeddings. 
In contrast, CNN based models learn more expressive embeddings due to their parameter efficiency 
and consideration of complex relations. 
However, both translational and CNN based models process each triple independently and hence fail to encapsulate the semantically rich and latent relations that are inherently present in the vicinity of a given entity in a KG.

Motivated by the aforementioned observations, we propose a \emph{generalized} attention-based graph embedding for relation prediction.
For node classification, \emph{graph attention networks} (GATs)~\cite{velickovic2018graph} have been shown to focus on the most relevant portions of the graph, namely the node features in a 1-hop neighborhood. Given a KG and the task of relation prediction, our model \emph{generalizes} and \emph{extends} the attention mechanism by guiding attention to both entity (node) and relation (edge) features in a multi-hop neighborhood of a given entity / node.

Our idea is: 1) to capture multi-hop relations~\cite{Lin2015} surrounding a given node, 2) to encapsulate the diversity of roles played by an entity in various relations, and 3) to consolidate the existing knowledge present in semantically similar relation clusters~\cite{Valverde_2012}. 
Our model achieves these objectives by assigning different weight mass (attention) to nodes in a neighborhood and by propagating attention via layers in an iterative fashion. However, as the model depth increases, the contribution of distant entities decreases exponentially. To resolve this issue, we use relation composition as proposed by \cite{Lin2015} to introduce an auxiliary edge between $n$-hop neighbors, which then readily allows the flow of knowledge between entities. 
Our architecture is an encoder-decoder model where our \emph{generalized graph attention model} and \emph{ConvKB}~\cite{nguyen2018novel} play the roles of an \emph{encoder} and \emph{decoder}, respectively. Moreover, this method can be extended for learning effective embeddings for Textual Entailment Graphs ~\cite{kotlerman_dagan_magnini_bentivogli_2015}, where global learning has proven effective in the past as shown by ~\cite{berant-etal-2015-efficient} and ~\cite{berant-etal-2010-global}.

Our contributions are as follows.
To the best of our knowledge, we are the first to learn new graph attention based embeddings that specifically target relation prediction on KGs. Secondly, we generalize and extend graph attention mechanisms to capture both entity and relation features in a multi-hop neighborhood of a given entity. Finally, we evaluate our model on challenging relation prediction tasks for a wide variety of real-world datasets. Our experimental results indicate a clear and substantial improvement over state-of-the-art relation prediction methods.
For instance, our attention-based embedding achieves an improvement of $104 \%$ over the state-of-the-art method for the Hits@$1$ metric on the popular \emph{Freebase (FB15K-237)} dataset. 

The rest of the paper is structured as follows. We first provide a review of related work in Section~\ref{related_work} and then our detailed approach in Section~\ref{our_approach}. Experimental results and dataset descriptions are reported in 
Section~\ref{experiments} followed by our conclusion and future research directions in Section~\ref{conclusion}.

\section{Related Work}\label{related_work}
Recently, several variants of KG embeddings have been proposed for relation prediction.
These methods can be broadly classified as: (i) compositional, (ii) translational, 
(iii) CNN based, and (iv) graph based models. 

RESCAL~\cite{Nickel2011}, NTN~\cite{socher2013reasoning}, and the Holographic embedding
model (HOLE)~\cite{nickel2016holographic} are examples of compositional based models. 
Both RESCAL and NTN use tensor products which capture rich interactions, but 
require a large number of parameters to model relations and are thus cumbersome to compute. 
To combat these drawbacks, HOLE creates more efficient and scalable 
compositional representations using the circular correlation of entity embeddings.


In comparison, translational models like TransE~\cite{NIPS2013_5071}, DISTMULT~\cite{yang2014} and ComplEx~\cite{trouillon2016complex} propose arguably simpler models. 
TransE considers the translation operation between head and tail entities for relations. 
DISTMULT~\cite{yang2014} learns embeddings using a \emph{bilinear diagonal model} which is a special case of the bilinear objective used in NTN and TransE. DISTMULT uses weighted element-wise dot products to model entity relations.
ComplEx~\cite{trouillon2016complex} generalizes DISTMULT~\cite{yang2014} by using complex  
embeddings and Hermitian dot products instead. 
These translational models are faster, require fewer parameters and are relatively easier to train, but 
result in less expressive KG embeddings.

Recently, two CNN based models have been proposed for relation prediction, namely ConvE~\cite{dettmers2018convolutional} and ConvKB~\cite{nguyen2018novel}. 
ConvE uses 2-D convolution over embeddings to
predict links. It comprises of a convolutional layer, a fully connected projection layer and an inner product layer for the final predictions. 
Different feature maps are generated using multiple filters to extract global relationships.
Concatenation of these feature maps represents an input triple. 
These models are parameter efficient but 
consider each triple independently without taking into account the relationships between the triples.

A graph based neural network model called R-GCN~\cite{schlichtkrull2018modeling} is an extension of applying \emph{graph convolutional networks} (GCNs)~\cite{kipf2017semi} to relational data. 
It applies a convolution operation to the neighborhood of each entity and assigns them equal weights. 
This graph based model does not outperform the CNN based models.

Existing methods either learn KG embeddings by solely focusing on entity features or by taking into account the features of entities and relations in a disjoint manner. Instead, our proposed graph attention model holistically captures multi-hop and semantically similar relations in the $n$-hop neighborhood of any given entity in the KG.


\section{Our Approach}\label{our_approach}
We begin this section by introducing the notations and definitions used in the rest of the paper, 
followed by a brief background on \emph{graph attention networks} (GATs)~\cite{velickovic2018graph}. Finally, we describe our proposed attention architecture for knowledge graphs followed by our decoder network. 

\subsection{Background}
\label{ssec:background}
A knowledge graph is denoted by $\mathcal{G}=(\mathcal{E}, R)$, where $\mathcal{E}$ and $R$ represent the set of entities (nodes) and relations (edges), respectively. A triple \((e_s, r, e_o)\) is represented as an edge $r$ between nodes $e_s$ and $e_r$ in $\mathcal{G}$\footnote{From here onwards, the pairs ``\emph{node} / \emph{entity}" and ``\emph{edge} / \emph{relation}" will be used interchangeably}.
Embedding models try to learn an effective representation of entities, relations, and a scoring function \(f\), such that for a given input triple \(t = (e_s, r, e_o)\), \(f(t)\) gives the likelihood of \(t\) being a valid triple. For example, 
Figure~\ref{fig:toys} shows the subgraph from a KG which infers missing links represented by dashed lines using existing triples such as \textit{(London, captial\_of, United Kingdom)}.

\subsection{Graph Attention Networks (GATs)}
\label{ssec:background}
\emph{Graph convolutional networks} (GCNs) \cite{kipf2017semi} gather information from the entity's neighborhood and all  neighbors contribute equally in the information passing.
To address the shortcomings of GCNs, \cite{velickovic2018graph} introduced \emph{graph attention networks} (GATs). GATs learn to assign varying levels of importance to nodes in every node's neighborhood, rather than treating all neighboring nodes with equal importance, as is done in GCN. 

The input feature set of nodes to a layer is \(\textbf{x} = \{\vec{x}_{1},\vec{x}_{2},...,\vec{x}_{N}\}\). A layer produces a transformed set of node feature vectors \(\textbf{x}^\prime = \{\vec{x}_{1}^{\prime},\vec{x}_{2}^{\prime},...,\vec{x}_{N}^{\prime}\}\), where \(\vec{x}_{i}\) and \(\vec{x}_{i}^{\prime}\) are input and output embeddings of the entity \(e_i\), and \(N\) is number of entities (nodes). A single GAT layer can be described as
\begin{equation}\label{eq:eij}
 e_{ij} = a(\textbf{W} \vec{x_{i}},\textbf{W} \vec{x_{j}})
\end{equation}
where \(e_{ij}\) is the attention value of the edge \((e_i, e_j)\) in $\mathcal{G}$, \(\textbf{W}\) is a parametrized linear transformation matrix mapping the input features to a higher dimensional output feature space, and \(a\) is any \emph{attention function} of our choosing.

Attention values for each edge are the \textit{importance} of the edge \((e_i, e_j)^{'}\)s features for a source node \(e_i\). Here, the relative attention \(\alpha_{ij}\) is computed using a \emph{softmax function} over all the values in the neighborhood.
Equation \ref{eq:hiprime} shows the output of a layer. GAT employs \emph{multi-head attention} to stabilize the learning process as credited to \cite{NIPS2017_7181}. 
\begin{equation}\label{eq:hiprime}
\vec{x_{i}^{\prime}} = \sigma \Bigg( \sum_{j \in \mathcal{N}_{i}} \alpha_{ij} \textbf{W} \vec{x_{j}} \Bigg)
\end{equation}
The multihead attention process of concatenating $K$ attention heads is shown as follows in Equation \ref{eq:hiconcat}. 
\begin{equation}\label{eq:hiconcat}
 \vec{x_{i}^{\prime}} = \underset{k=1}{\stackrel{K}{\Big \Vert}}  \sigma \Bigg( \sum_{j \in \mathcal{N}_{i}} \alpha_{ij}^{k} \textbf{W}^{k} \vec{x_{j}} \Bigg)
\end{equation}
where $\Vert$ represents concatenation, \(\sigma\) represents any non-linear function, $\alpha_{ij}^{k}$ are normalized attention coefficients of edge \((e_i, e_j)\) calculated by the $k$-th attention mechanism, and \(\textbf{W}^k\) represents the corresponding linear transformation matrix of the \(k\)-th 
attention mechanism.
The output embedding in the final layer is calculated using \emph{averaging}, instead of the concatenation operation, to achieve multi-head attention, as is shown in the following Equation \ref{eq:hisummation}.
\begin{equation}\label{eq:hisummation}
 \vec{x_{i}^{\prime}} =   \sigma \Bigg(\frac{1}{K}\sum_{k = 1}^{K}\sum_{j \in \mathcal{N}_{i}} \alpha_{ij}^{k} \textbf{W}^{k} \vec{x_{j}} \Bigg)
\end{equation}

\subsection{Relations are Important}
Despite the success of GATs, they are unsuitable for KGs as they ignore relation (edge) features, which are an 
integral part of KGs. In KGs, entities play different roles depending on the \emph{relation} they are associated with.
For example, in Figure \ref{fig:toys}, entity \textit{Christopher Nolan} appears in two different triples assuming the roles of a
\emph{brother} and a \emph{director}. To this end, we propose a novel embedding approach to incorporate \textit{relation and neighboring node features} in the attention mechanism.
\begin{figure}
  \includegraphics[width=\linewidth]{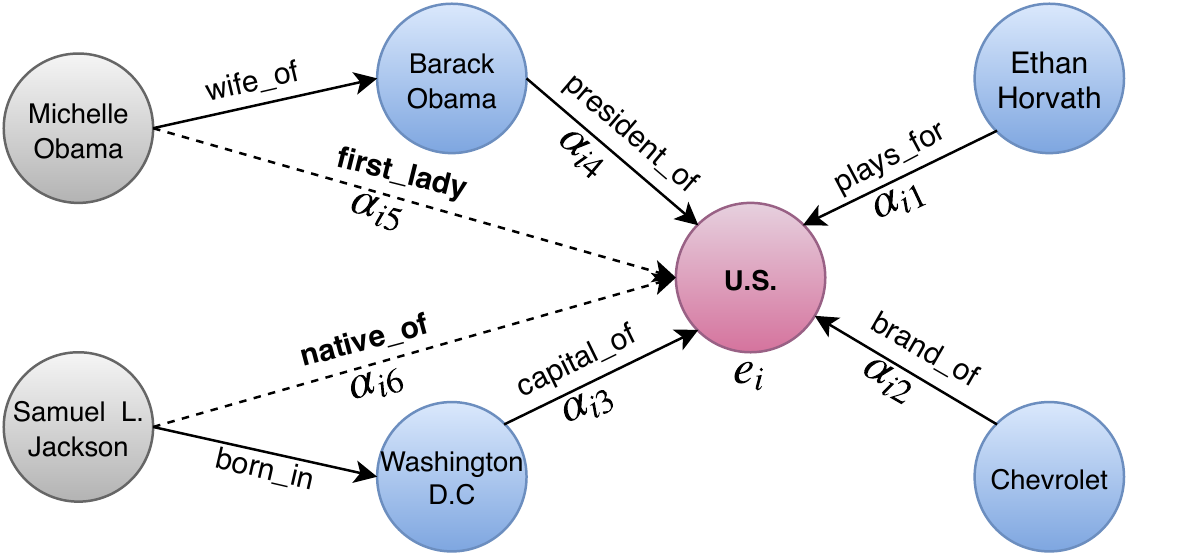}
  \caption{This figure shows the aggregation process of our graph attentional layer. \(\alpha_{ij}\) represents relative attention values of the edge. 
  	The dashed lines represent an \textit{auxiliary} edge from a $n$-hop neighbors, in this case $n = 2$. }
  \label{fig:attentionex}
\end{figure}
We define a single attentional layer, which is the building block of our model. Similar to GAT, our framework is agnostic to the particular choice of attention mechanism.

Each layer in our model takes two embedding matrices as input. \emph{Entity embeddings} are represented by a matrix \(\textbf{H} \in \mathbb{R}^{N_e \times T} \), where the \(i\)-th row is the embedding of entity  \(e_i\), \(N_e\) is the total number of entities, and \(T\) is the feature dimension of each entity embedding. 
With a similar construction, the \emph{relation embeddings} are represented by a matrix \(\textbf{G} \in \mathbb{R}^{N_r \times P}\). The layer then outputs the corresponding embedding matrices, \(\textbf{H}^\prime \in \mathbb{R}^{N_e \times T^\prime} \) and  \(\textbf{G}^\prime \in \mathbb{R}^{N_r \times P^\prime} \).

\begin{figure*}
  \includegraphics[width=\linewidth]{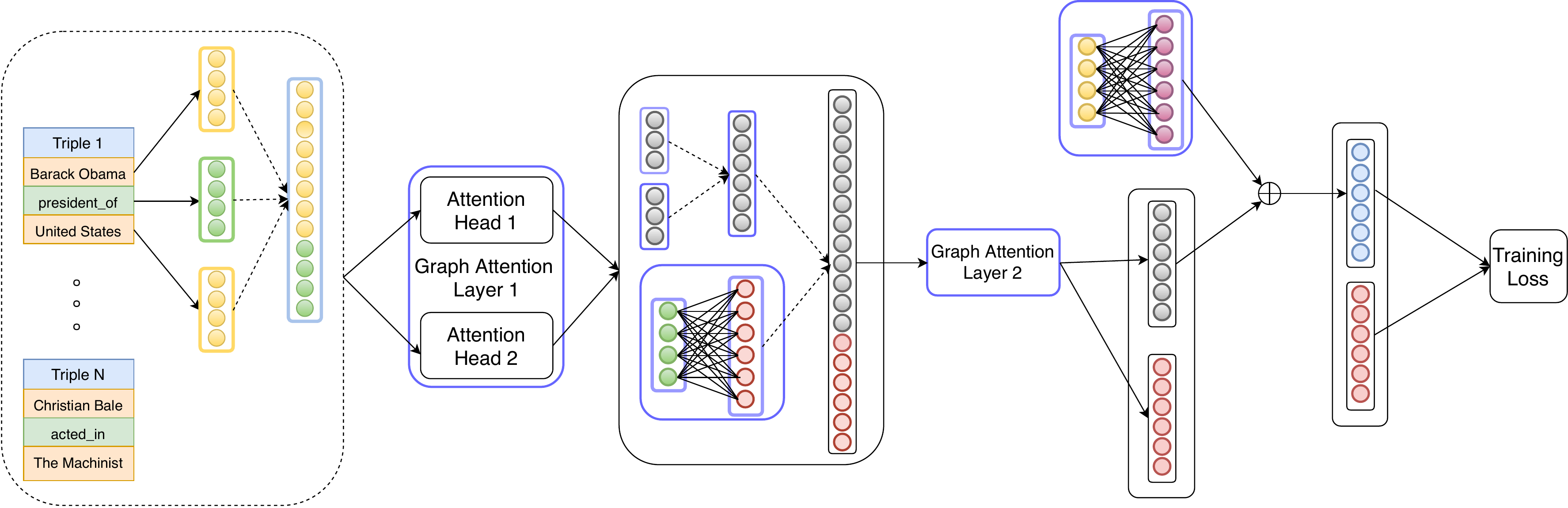}
  \caption{This figure shows end-to-end architecture of our model. Dashed arrows in the figure represent concatenation operation. Green circles represents initial entity embedding vectors and yellow circles represents initial relation embedding vectors.}
  \label{fig:architecture}
\end{figure*}

In order to obtain the new embedding for an entity \(e_i\), a representation of each triple associated with \(e_i\) is learned. We learn these embeddings by performing a linear transformation over the concatenation of entity and relation feature vectors corresponding to a particular triple \(t_{ij}^{k} = (e_i, r_k, e_j)\), as is shown in Equation \ref{eq:cijk}. This operation is also illustrated in the initial block of Figure~\ref{fig:architecture}.
\begin{equation}\label{eq:cijk}
 \vec{c_{ijk}} = \textbf{W}_{1} [\vec{h}_{i} \Vert \vec{h}_{j} \Vert \vec{g}_{k}] 
\end{equation}
where \(\vec{c_{ijk} }\) is the vector representation of a triple \(t_{ij}^k\). Vectors \(\vec{h}_i, \vec{h}_j\), and \(\vec{g}_k\) denote embeddings of entities \(e_i, e_j\) and relation \(r_k\), respectively. Additionally, \(\textbf{W}_1\) denotes the linear transformation matrix. 
\begin{figure}[tbp]
\centering
  \includegraphics[trim={0 0.8cm 1.4cm 0},clip,width=0.9\linewidth, height=0.9
\linewidth, angle = 90]{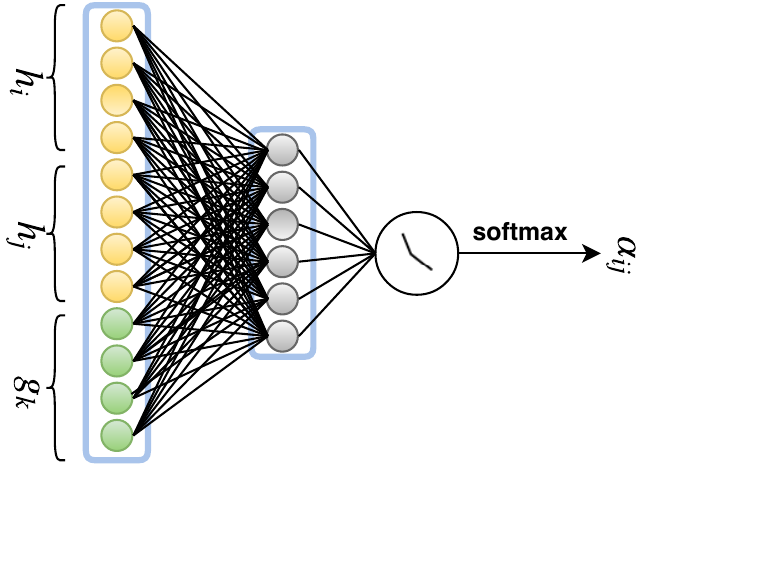}
  \caption{Attention Mechanism}
  \label{fig:layer}
\end{figure}
Similar to \cite{velickovic2018graph}, we learn the \emph{importance} of each triple \(t_{ij}^{k}\) denoted by \(b_{ijk}\). 
We perform a linear transformation parameterized by a weight matrix \(\textbf{W}_2\) followed by application of 
the LeakyRelu non-linearity to get the absolute attention value of the triple (Equation \ref{eq:bijk}).
\begin{equation}\label{eq:bijk}
 b_{ijk} = \textrm{LeakyReLU} \Big( \textbf{W}_{2} c_{ijk} \Big)
\end{equation}
To get the relative attention values \emph{softmax} is applied over \(b_{ijk}\) as shown in Equation \ref{eq:alphaijk}. Figure \ref{fig:layer} shows the computation of relative attention values \(\alpha_{ijk}\) for a single triple.
\begin{align}
\label{eq:alphaijk}
 \alpha_{ijk} &= \textrm{softmax}_{jk} (b_{ijk}) \\ \nonumber
 &= \frac{\textrm{exp} (b_{ijk})}{\sum_{n \in \mathcal{N}_{i}} \sum_{r \in \mathcal{R}_{in}} \textrm{exp} (b_{inr})}
\end{align}
where \(\mathcal{N}_i\) denotes the neighborhood of entity \(e_i\) and \(\mathcal{R}_{ij}\) denotes the set of relations connecting entities \(e_i\) and \(e_j\).
The new embedding of the entity \(e_i\) is the sum of each triple representation weighted by their attention values as shown in Equation \ref{eq:hiprimeKBGAT}.
\begin{equation}\label{eq:hiprimeKBGAT}
  \vec{h_{i}^{\prime}} = \sigma \Bigg( \sum_{j \in \mathcal{N}_{i}} \sum_{k \in \mathcal{R}_{ij}} \alpha_{ijk} \vec{c_{ijk}} \Bigg)
\end{equation}
As suggested by \cite{velickovic2018graph}, multi-head attention which was first introduced by \cite{NIPS2017_7181}, is used to stabilize the learning process and encapsulate more information about the neighborhood. Essentially, \(M\) independent attention mechanisms calculate the embeddings, which are then concatenated, resulting in the following representation: 
\begin{equation}\label{eq:sumhiprime}
 \vec{h_{i}^{\prime}} = \underset{m=1}{\stackrel{M}{\Big \Vert}}  \sigma \Bigg( \sum_{j \in \mathcal{N}_{i}} \alpha_{ijk}^{m} c_{ijk}^{m} \Bigg)
\end{equation}
This is the \emph{graph attention layer} shown in Figure ~\ref{fig:architecture}.
We perform a linear transformation on input \emph{relation embedding} matrix \(\textbf{G}\), parameterized by a weight matrix \(\textbf{W}^R \in \mathbb{R}^{T \times T^\prime}\), where \(T^\prime\) is the dimensionality of output \emph{relation embeddings} (Equation~\ref{eq:Rprime}).
\begin{equation}\label{eq:Rprime}
 G^{\prime} = G.\textbf{W}^{R}
\end{equation}
In the final layer of our model, instead of concatenating the embeddings from multiple heads we employ averaging to get final embedding vectors for entities as shown in Equation \ref{eq:hisummationKB}.
\begin{equation}\label{eq:hisummationKB}
 \vec{h_{i}^{\prime}} =   \sigma \Bigg(\frac{1}{M}\sum_{m = 1}^{M}\sum_{j \in \mathcal{N}_{i}}\sum_{k \in \mathcal{R}_{ij}} \alpha_{ijk}^{m}c_{ijk}^{m} \Bigg)
\end{equation}
However, while learning new embeddings, entities lose their initial embedding information. To resolve this issue, we linearly transform \(\textbf{H}^i\) to obtain \(\textbf{H}^t\) using a weight matrix \(\textbf{W}^E \in \mathbb{R}^{T^i \times T^f}\), where \(\textbf{H}^i\) represents the input entity embeddings to our model, \(\textbf{H}^t\) represents the transformed entity embeddings, \(T^i\) denotes the dimension of an initial entity embedding, and \(T^f\) denotes the dimension of the final entity embedding. We add this initial entity embedding information to the entity embeddings obtained from the final attentional layer, \(\textbf{H}^f \in \mathbb{R}^{N_e \times T^f}\) as shown in Equation \ref{eq:final}.
\begin{equation}\label{eq:final}
 \textbf{H}^{\prime\prime} = \textbf{W}^E \textbf{H}^t + \textbf{H}^{f}
\end{equation}
In our architecture, we extend the notion of an \emph{edge} to a \emph{directed path}
by introducing an auxiliary relation for $n$-hop neighbors between two entities. 
The embedding of this auxiliary relation is the summation of embeddings of all the relations in the path.
Our model iteratively accumulates knowledge from distant neighbors of an entity. As illustrated in figure \ref{fig:attentionex}, in the first layer of our model, all entities capture information from their \emph{direct in-flowing neighbors}. In the second layer, \textit{U.S} gathers information from entities \textit{Barack Obama, Ethan Horvath, Chevrolet, and Washington D.C}, which already possess information about their neighbors \textit{Michelle Obama} and \textit{Samuel L. Jackson}, from a previous layer. In general, for a $n$ layer model the incoming information is accumulated over a $n$-hop neighborhood. The aggregation process to learn new entity embeddings and the introduction of an auxiliary edge between $n$-hop neighbors is also shown in Figure \ref{fig:attentionex}. We normalize the entity embeddings after every generalized GAT layer and prior to the first layer, for every main iteration.

\subsection{Training Objective}
Our model borrows the idea of a \emph{translational scoring function} from \cite{NIPS2013_5071}, which learns embeddings such that for a given \emph{valid} triple \(t_{ij}^{k} = (e_i, r_k, e_j)\), the condition  \(\vec{h}_i + \vec{g}_k \approx \vec{h}_j\) holds, i.e., \(e_j\) is the nearest neighbor of \(e_i\) connected via relation \(r_k\). 
Specifically, we try to learn \emph{entity} and \emph{relation} embeddings to minimize the L1-norm dissimilarity measure given by $d_{t_{ij}} = \lVert \vec{h_i} + \vec{g_k} - \vec{h_j} \rVert_{1}$. 

We train our model using \emph{hinge-loss} which is given by the following expression
\begin{equation}\label{eq:hingeloss}
L(\Omega) = \sum_{t_{ij} \in S} \sum_{t_{ij}^{\prime} \in S^{\prime}} \max\{d_{t_{ij}^{\prime}} - d_{t_{ij}} + \gamma,0\}
\end{equation}
where \(\gamma > 0\) is a margin hyper-parameter, $S$ is the set of valid triples, and $S'$ denotes the set of invalid triples, given formally as 
\begin{equation}\label{eq:invalid}\nonumber
  S^{\prime} = 
  \underbrace{\{t_{i^{\prime}j}^{k} \mid e_i^{\prime} \in \mathcal{E} \setminus e_i \}}_{\text{replace head entity} }
  \cup 
  \underbrace{\{t_{ij^{\prime}}^{k} \mid e_j^{\prime} \in \mathcal{E} \setminus e_j\}}_{\text{replace tail entity} }
\end{equation}
\begin{table*}[t!]
	\centering\small
	\begin{tabular}{l|cccccccc}
		\hline
		&               &              & \multicolumn{4}{c}{ \textbf{\# Edges}} \\ 
		\cline{4-7}
		\textbf{Dataset}   &  \textbf{\# Entities}  & \textbf{\# Relations}  & Training  & Validation  & Test & Total & \textbf{Mean in-degree} & \textbf{Median in-degree}\\ 
		\hline
		\emph{WN18RR}    & 40,943 & 11  & 86,835   & 3034   & 3134   & 93,003  & 2.12  & 1 \\ 
		\emph{FB15k-237} & 14,541 & 237 & 272,115  & 17,535 & 20,466 & 310,116 & 18.71 & 8  \\ 
		\emph{NELL-995}  & 75,492 & 200 & 149,678  & 543    & 3992   & 154,213 & 1.98  & 0   \\
		\emph{Kinship}   & 104    & 25  & 8544     & 1068   & 1074   & 10,686  & 82.15 & 82.5 \\
		\emph{UMLS}      & 135    & 46  & 5216     & 652    & 661    & 6529    & 38.63 & 20    \\
		\hline
	\end{tabular}
	\caption{Dataset statistics}
	\label{tb:datasets}
\end{table*}
\subsection{Decoder}
Our model uses ConvKB \cite{nguyen2018novel} as a \emph{decoder}. 
The aim of the convolutional layer is to analyze the global embedding properties of a triple $t_{ij}^{k}$ across each dimension and to generalize the transitional characteristics in our model. 
The score function with multiple feature maps can be written formally as:
\begin{equation*}\label{eq:sumhiprime}
 f(t_{ij}^{k}) = \Bigg (\underset{m=1}{\stackrel{\Omega}{\Big \Vert}}  \textrm{ReLU}([\vec{h}_i, \vec{g}_k, \vec{h}_j]*\omega^m)\Bigg).\textbf{W}
\end{equation*}
where \(\omega^m\) represents the \(m^{th}\) convolutional filter, \(\Omega\) is a hyper-parameter denoting number of filters used, $*$ is a convolution operator, and \(\textbf{W} \in \mathbb{R}^{\Omega k \times 1}\) represents a linear transformation matrix used to compute the final score of the triple. 
The model is trained using soft-margin loss as 
\begin{equation*}\label{eq:soft-margin}
 \mathcal{L} = \sum_{t_{ij}^{k} \in \{S \cup S^{\prime}\}} \textrm{log} (1 + \textrm{exp} (l_{t_{ij}^{k}}. f(t_{ij}^{k})) ) + \frac{\lambda}{2} \Vert \textbf{W} \Vert_{2}^{2}
\end{equation*}
where $l_{t_{ij}^{k}} = \begin{cases}
1 &\text{for $t_{ij}^{k} \in S$}\\
-1 &\text{for $t_{ij}^{k} \in S^{\prime}$}
\end{cases}$

\section{Experiments and Results}\label{experiments}

\begin{table*}[t!]
	\centering\small
	\setlength{\tabcolsep}{4pt} 
	\renewcommand{\arraystretch}{1} 
	\begin{tabular}{lccccclccccc}
		\hline
		& \multicolumn{5}{c}{\textbf{WN18RR}} &  & \multicolumn{5}{c}{\textbf{FB15K-237}}         \\ \cline{2-6} \cline{8-12} 
		
		&  &  & \multicolumn{3}{c}{Hits@N} &  &  &  & \multicolumn{3}{c}{Hits@N} \\ \cline{4-6} \cline{10-12} 
		
		& MR & MRR & @1 & @3 & @10 &    & MR & MRR & @1 & @3 & @10    \\
		\hline
		\\
		DistMult \cite{yang2014} & 7000 & 0.444 & \underline{41.2} & \underline{47} & 50.4 &    & 512 & 0.281 & 19.9 & 30.1 & 44.6  \\
		
		ComplEx \cite{trouillon2016complex} & 7882 & \underline{0.449} & 40.9 & 46.9 & 53 &    & 546 & 0.278 & 19.4 & 29.7 & 45 \\
		
		ConvE \cite{dettmers2018convolutional} & 4464 & \textbf{0.456} & \textbf{41.9} & \underline{47} & 53.1 &    & 245 & \underline{0.312} & \underline{22.5} & 34.1 & \underline{49.7} \\
		
		TransE \cite{NIPS2013_5071} & 2300 & 0.243 & 4.27 & 44.1 & 53.2 &    & 323 & 0.279 & 19.8     & \underline{37.6} & 44.1    \\
		
		ConvKB \cite{nguyen2018novel} & \textbf{1295} & 0.265 & 5.82 & 44.5 & \underline{55.8} &    & \underline{216} & 0.289 & 19.8 & 32.4 & 47.1   \\
		
		R-GCN \cite{schlichtkrull2018modeling} & 6700 & 0.123 & 20.7 & 13.7 & 8 &    & 600 & 0.164 & 10 & 18.1 & 30 
		\\
		
		\hline
		Our work & \underline{1940} & 0.440 & 36.1 & \textbf{48.3} & \textbf{58.1} &    & \textbf{210} & \textbf{0.518} & \textbf{46} & \textbf{54} & \textbf{62.6}   \\
		
		\hline
	\end{tabular}
	\caption{Experimental results on WN18RR and FB15K-237 test sets. Hits@N values are in percentage. The best score is in \textbf{bold} and second best score is \underline{underlined}.
	}\label{tb:results1}
\end{table*}

\begin{table*}[t!]
\centering\small
\setlength{\tabcolsep}{4pt} 
\renewcommand{\arraystretch}{1} 
\begin{tabular}{lccccclccccc}
\hline
& \multicolumn{5}{c}{\textbf{NELL-995}} &    & \multicolumn{5}{c}{\textbf{Kinship}}    \\ 
\cline{2-6} \cline{8-12}  

&  &  & \multicolumn{3}{c}{Hits@N} &  &  &  & \multicolumn{3}{c}{Hits@N}    \\ 
\cline{4-6} \cline{10-12} 

& MR & MRR & @1 & @3 & @10 &    & MR & MRR & @1 & @3 & @10    \\
\hline
\\
DistMult \cite{yang2014} & 4213 & 0.485 & 40.1 & 52.4 & 61 &    & 5.26 & 0.516 & 36.7 & 58.1 & 86.7   \\

ComplEx \cite{trouillon2016complex} & 4600 & 0.482 & 39.9 & 52.8 & 60.6 &    & 2.48 & 0.823 & 73.3 & 89.9 & 97.11    \\

ConvE \cite{dettmers2018convolutional} & 3560 & \underline{0.491} & \underline{40.3} & \underline{53.1} & \underline{61.3} &    & \underline{2.03} & \underline{0.833} & \underline{73.8} & \underline{91.7} & \textbf{98.14}    \\

TransE \cite{NIPS2013_5071} & 2100 & 0.401 & 34.4 & 47.2 & 50.1 &    & 6.8 & 0.309 & 0.9 & 64.3 & 84.1 
\\

ConvKB \cite{nguyen2018novel} & \textbf{600} & 0.43 & 37.0 & 47 & 54.5 &    & 3.3 & 0.614 & 43.62 & 75.5 & 95.3 
   \\

R-GCN \cite{schlichtkrull2018modeling} & 7600 & 0.12 & 8.2 & 12.6 & 18.8 &    & 25.92 & 0.109 & 3 & 8.8 & 23.9 
\\
   
\hline
Our work & \underline{965} & \textbf{0.530} & \textbf{44.7} & \textbf{56.4} & \textbf{69.5} &    & \textbf{1.94} & \textbf{0.904} & \textbf{85.9} & \textbf{94.1} & \underline{98}  \\

\hline
\end{tabular}
\caption{Experimental results on NELL-995 and Kinship test sets. Hits@N values are in percentage. The best score is in \textbf{bold} and second best score is \underline{underlined}.
  }\label{tb:results2}
\end{table*}

\subsection{Datasets}
\label{ssec:datasets}
To evaluate our proposed method, we use five benchmark datasets: \emph{WN18RR}~\cite{dettmers2018convolutional}, 
 \emph{FB15k-237}~\cite{toutanova2015representing}, \emph{NELL-995}~\cite{xiong2017}, \emph{Unified Medical 
 Language Systems (UMLS)}~\cite{Kok:2007:SPI:1273496.1273551} and \emph{Alyawarra Kinship}~\cite{LinRX2018}. 
Previous works~\cite{toutanova2015representing,dettmers2018convolutional} suggest that the task of relation prediction in 
\emph{WN18} and \emph{FB15K} suffers from the problem of \emph{inverse relations}, whereby one can achieve state-of-the-art results using a simple \emph{reversal rule} 
based model, as shown by \cite{dettmers2018convolutional}. Therefore, corresponding subset datasets
 \emph{WN18RR} and \emph{FB15k-237} were created to resolve the reversible relation problem in \emph{WN18} and \emph{FB15K}. We used the data splits provided by~\cite{nguyen2018novel}. Table~\ref{tb:datasets} provides statistics of all datasets used.

\subsection{Training Protocol}
\label{ssec:training}
We create two sets of invalid triples, each time replacing either the head or tail entity in a triple by an invalid entity. 
We randomly sample equal number of invalid triples from both the sets to ensure robust performance on detecting both head and tail entity. Entity and relation embeddings produced by TransE~\cite{NIPS2013_5071,nguyen2018novel} are used to initialize our embeddings.

We follow a two-step training procedure, i.e., we first train our generalized GAT to encode information about the graph entities and relations and then train a decoder model like ConvKB \cite{nguyen2018novel} to perform the relation prediction task. The original GAT update Equation~\ref{eq:hiconcat} only aggregates information passed from \(1\)-hop neighborhood, while our generalized GAT uses information from the  \(n\)-hop neighborhood. We use auxiliary relations to aggregate more information about the neighborhood in sparse graphs. We use Adam to optimize all the parameters with initial learning rate set at $0.001$. Both the entity and relation embeddings of the final layer are set to $200$. The optimal hyper-parameters set for each dataset are mentioned in our supplementary section.

\subsection{Evaluation Protocol}
\label{ssec:evaluation}
In the relation prediction task, the aim is to predict a triple \((e_i, r_k, e_j)\) with \(e_i\) or \(e_j\) missing, i.e., 
predict \(e_i\) given \((r_k, e_j)\) or predict \(e_j\) given \((e_i, r_k)\). 
We generate a set of $(N-1)$ corrupt triples \emph{for each entity} \(e_i\) by replacing it with every other entity \({e_i}^{\prime} \in \mathcal{E} \setminus e_i \), then we assign a score to each such triple. Subsequently, we sort these scores in ascending order and get the rank of a correct triple \((e_i, r_k, e_j)\). Similar to previous work (\cite{NIPS2013_5071}, \cite{nguyen2018novel},  \cite{dettmers2018convolutional}), we evaluate all the models in a \emph{filtered} setting, i.e, during ranking we remove corrupt triples which are already present in one of the training, validation, or test sets. 
This whole process is repeated by replacing the tail entity \(e_j\), and averaged metrics are reported. We report mean reciprocal rank (MRR), mean rank (MR) and the proportion of correct entities in the top $N$ ranks (Hits@N) for $N = 1, 3,$ and $10$.

\subsection{Results and Analysis}
\label{ssec:results}
Tables \ref{tb:results1} and \ref{tb:results2} present the prediction results on the test sets of all the datasets. 
The results clearly demonstrate that our proposed method\footnote{\href{https://github.com/deepakn97/relationPrediction}{Our work}} significantly outperforms state-of-the-art results on five metrics for \emph{FB15k-237}, and on two metrics for \emph{WN18RR}.  
We downloaded publicly available source codes to reproduce results of the state-of-the-art  methods\footnote{\href{https://github.com/datquocnguyen/STransE}{TransE}}\footnote{\href{https://github.com/TimDettmers/ConvE}{DistMult}}\footnote{\href{https://github.com/TimDettmers/ConvE}{ComplEx}}\footnote{\href{https://github.com/MichSchli/RelationPrediction}{R-GCN}}\footnote{\href{https://github.com/TimDettmers/ConvE}{ConvE}}\footnote{\href{https://github.com/daiquocnguyen/ConvKB}{ConvKB}} on all the datasets.
\begin{figure*}[t]
  \centering
  \hspace{-1.9cm}
  \begin{subfigure}[b]{0.2\linewidth}\label{hfb:a}
    \includegraphics[trim={0 0 1.5cm 0},clip,width=1.89\linewidth]{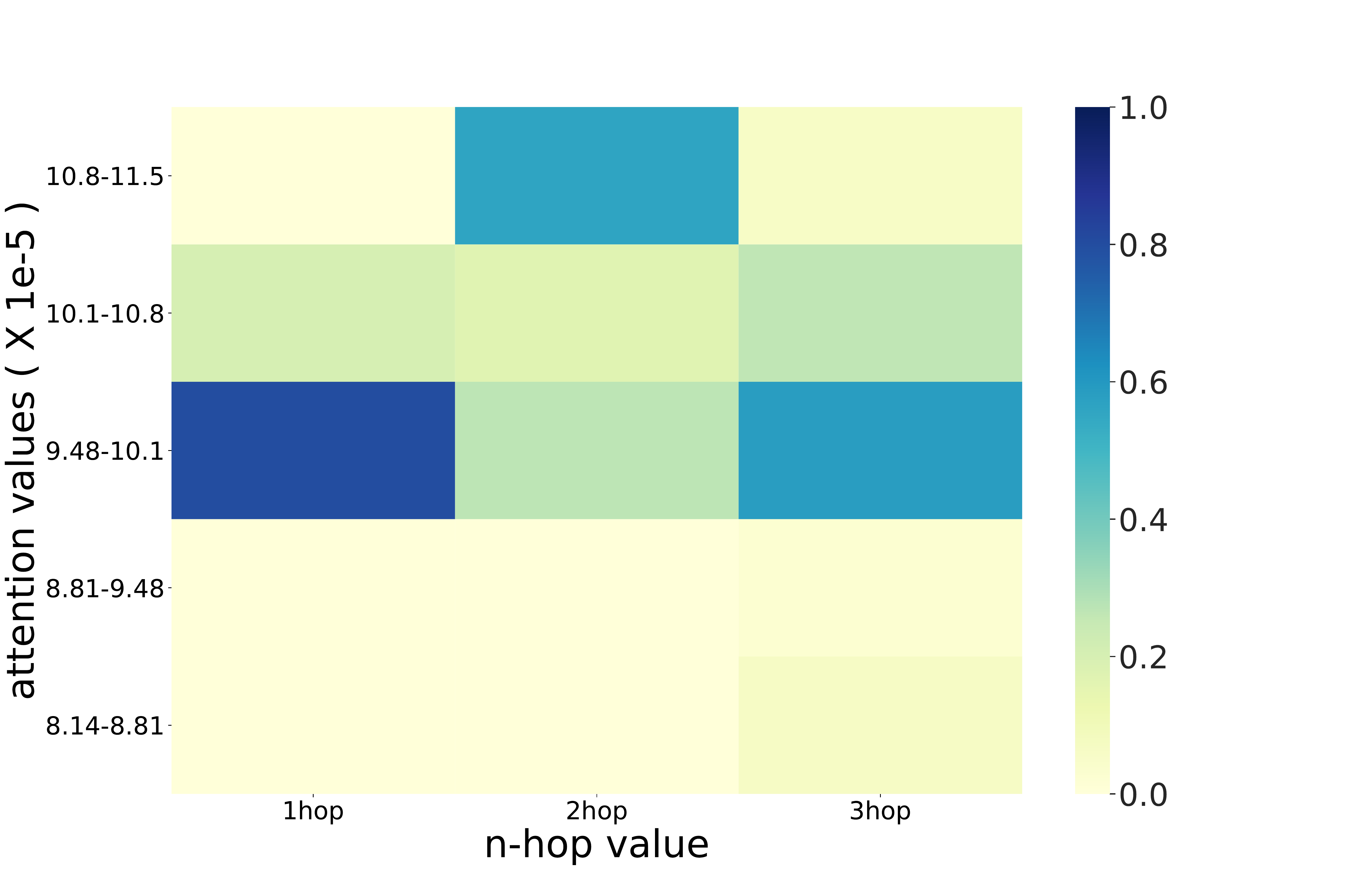}
    \caption{Epoch 0}
  \end{subfigure}
  \hspace{2.1cm}
  \begin{subfigure}[b]{0.2\linewidth}\label{hfb:c}
    \includegraphics[trim={0 0 1.5cm 0},clip,width=1.89\linewidth]{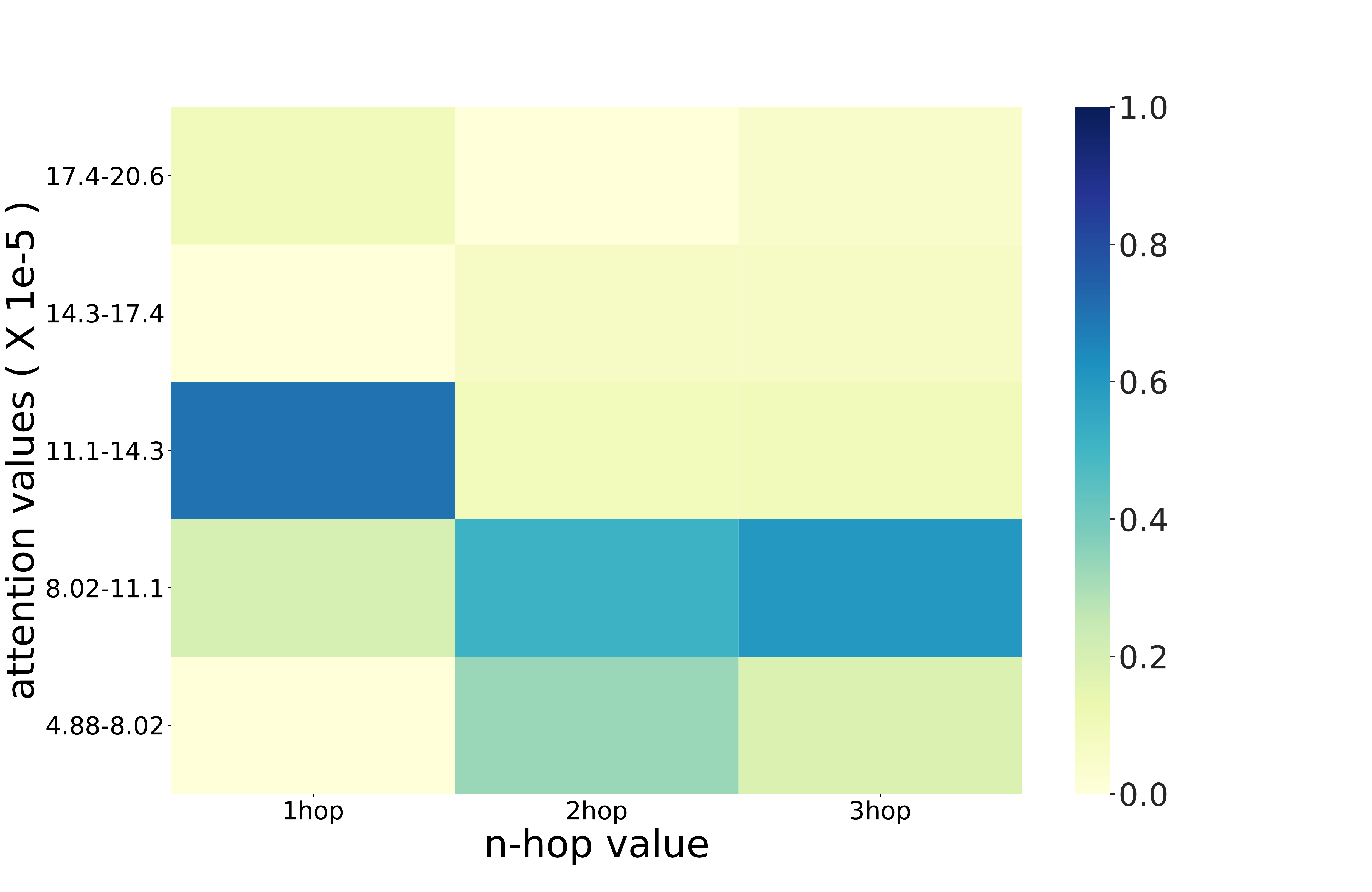}
    \caption{Epoch 1200}
  \end{subfigure}
  \hspace{2.1cm}
  \begin{subfigure}[b]{0.2\linewidth}\label{hfb:e}
    \includegraphics[trim={0 0 1.5cm 0},clip,width=1.89\linewidth]{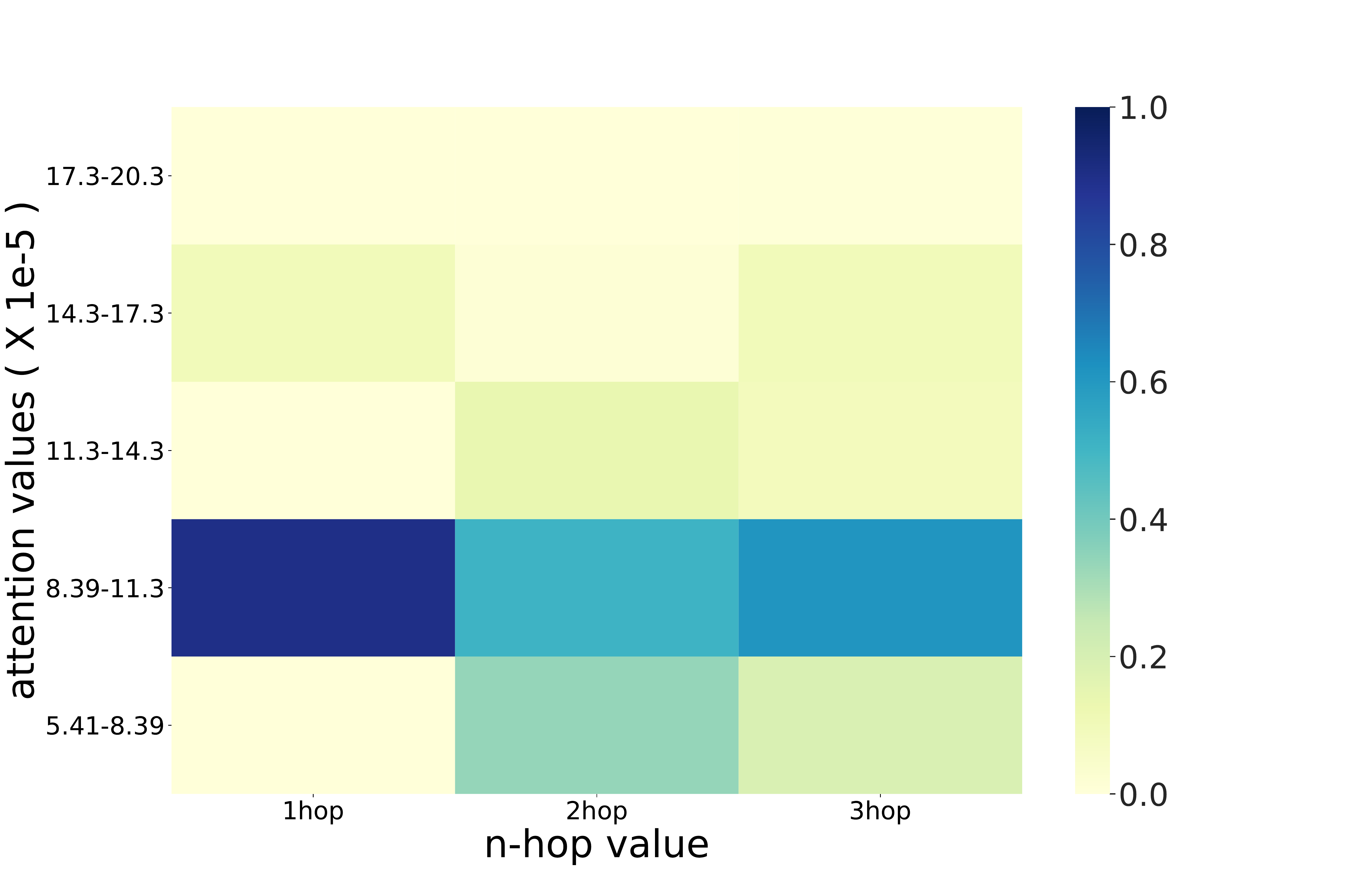}
    \caption{Epoch 2400}
  \end{subfigure}
  \caption{Learning process of our model on FB15K-237 dataset. Y-axis represents attention values \(\times 1e^{-5}\).}
  \label{fig:theat_fb}
\end{figure*}
\begin{figure*}[t]
  \centering
  \hspace{-1.9cm}
  \begin{subfigure}[b]{0.2\linewidth}\label{hwn:a}
    \includegraphics[trim={0 0 1.5cm 0},clip,width=1.89\linewidth]{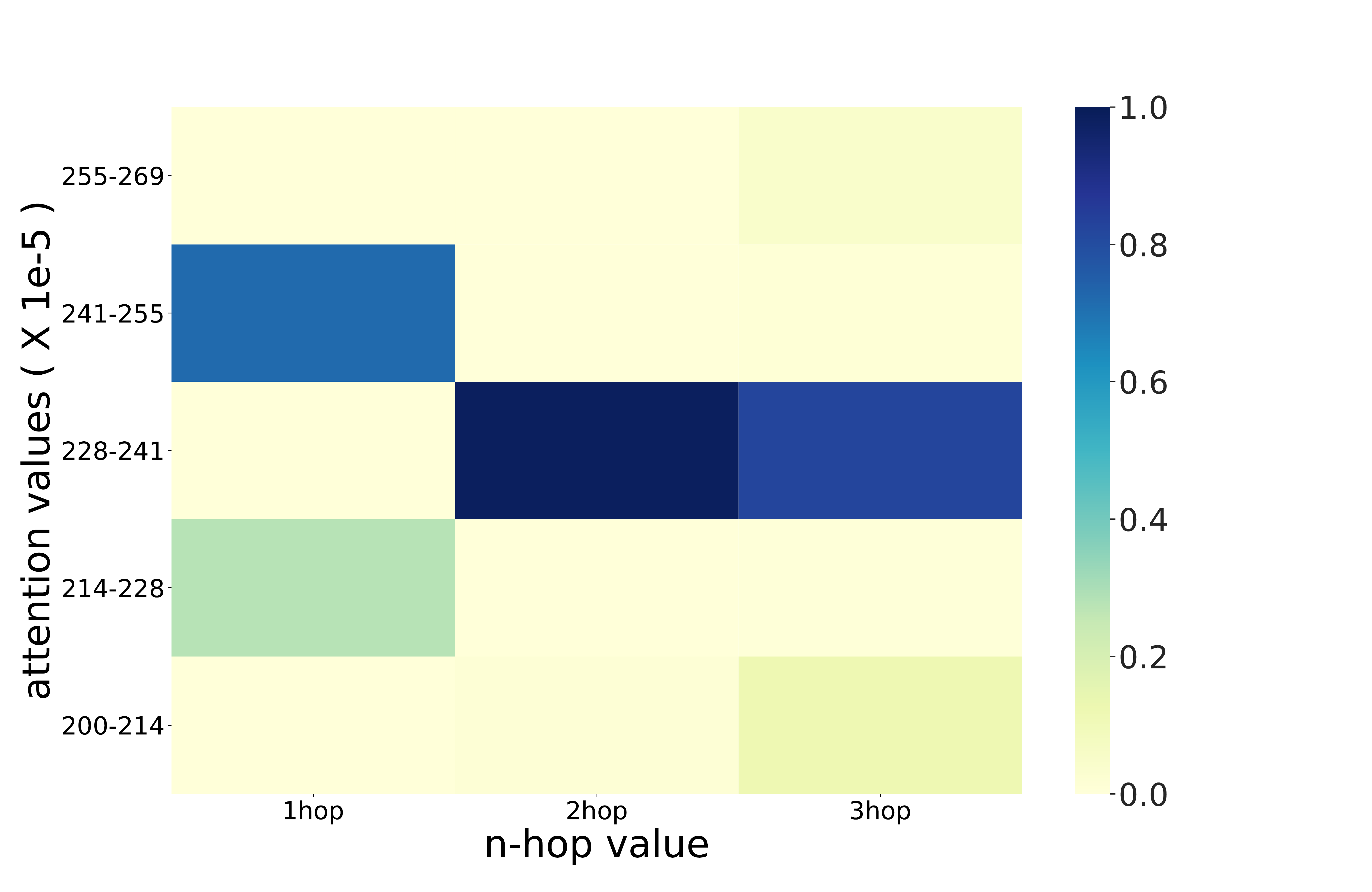}
    \caption{Epoch 0}
  \end{subfigure}
  \hspace{2.1cm}
  \begin{subfigure}[b]{0.2\linewidth}\label{hwn:c}
    \includegraphics[trim={0 0 1.5cm 0},clip,width=1.89\linewidth]{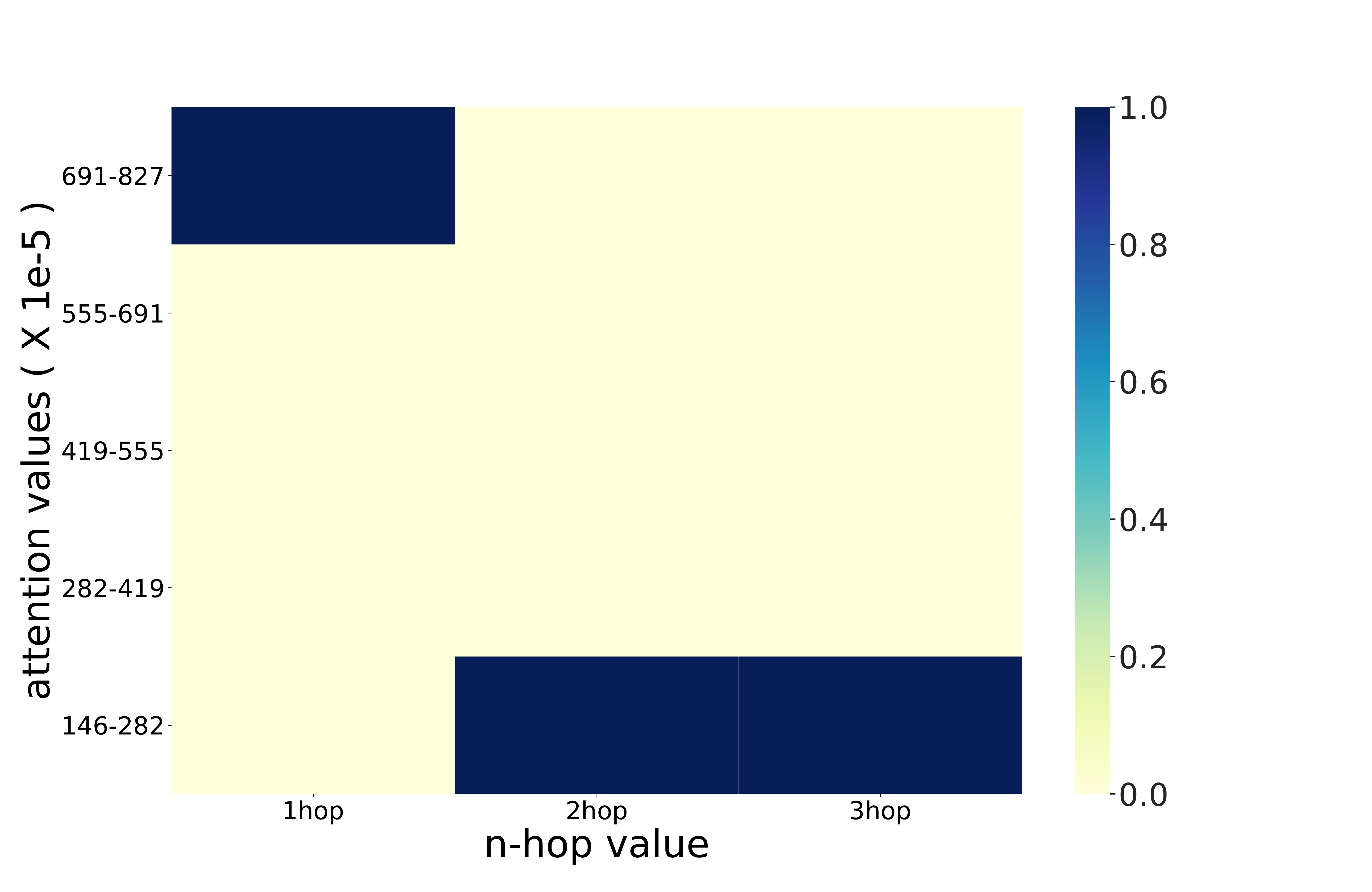}
    \caption{Epoch 1800}
  \end{subfigure}
  \hspace{2.1cm}
  \begin{subfigure}[b]{0.2\linewidth}\label{hwn:e}
    \includegraphics[trim={0 0 1.5cm 0},clip,width=1.89\linewidth]{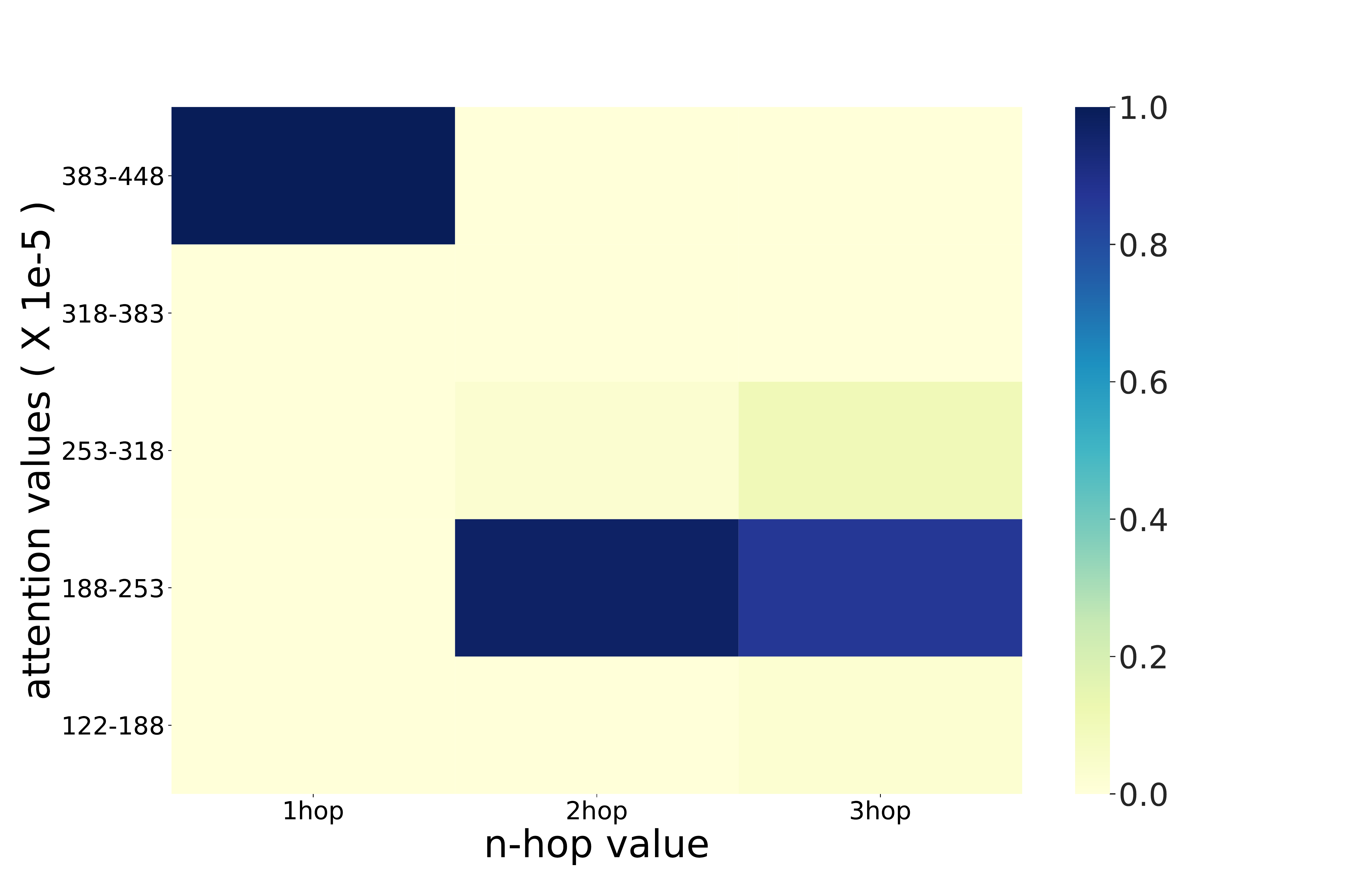}
    \caption{Epoch 3600}
  \end{subfigure}
  \caption{Learning process of our model on WN18RR dataset. Y-axis represents attention values \(\times 1e^{-5}\) }
  \label{fig:theat_wn}
\end{figure*}

\noindent\textbf{Attention Values vs Epochs}: We study the distribution of attention with increasing epochs for a particular node. 
Figure \ref{fig:theat_fb} shows this distribution on \emph{FB15k-237}. In the initial stages of the learning process, the attention is distributed randomly. As the training progresses and our model gathers more information from the neighborhood, it assigns more attention to direct neighbors and takes minor information from the more distant neighbors. Once the model converges, it learns to gather multi-hop and clustered relation information from the $n$-hop neighborhood of the node.
 
\noindent\textbf{PageRank Analysis}:
We hypothesize that complex and hidden multi-hop relations among entities are captured more succinctly in dense graphs as opposed to sparse graphs. 
To test this hypothesis, we perform an analysis similar to ConvE, where they study the \emph{correlation} between \emph{mean PageRank} and \emph{increase in MRR relative to DistMult}. 
We notice a strong correlation coefficient of 
\(r = 0.808\). Table \ref{tb:ablation} indicates that when there is an increase in PageRank values, there is also a corresponding increase in MRR values.
We observe an anomaly to our observed correlation in case of \emph{NELL-995} versus \emph{WN18RR}
and attribute this to the highly sparse and \emph{hierarchical structure} of \emph{WN18RR} which poses as a challenge to our method that does not capture information in a top-down recursive fashion.

\subsection{Ablation Study}
\label{sec:ablation}
\begin{figure}[h]
\centering
  \includegraphics[width=0.8\linewidth]{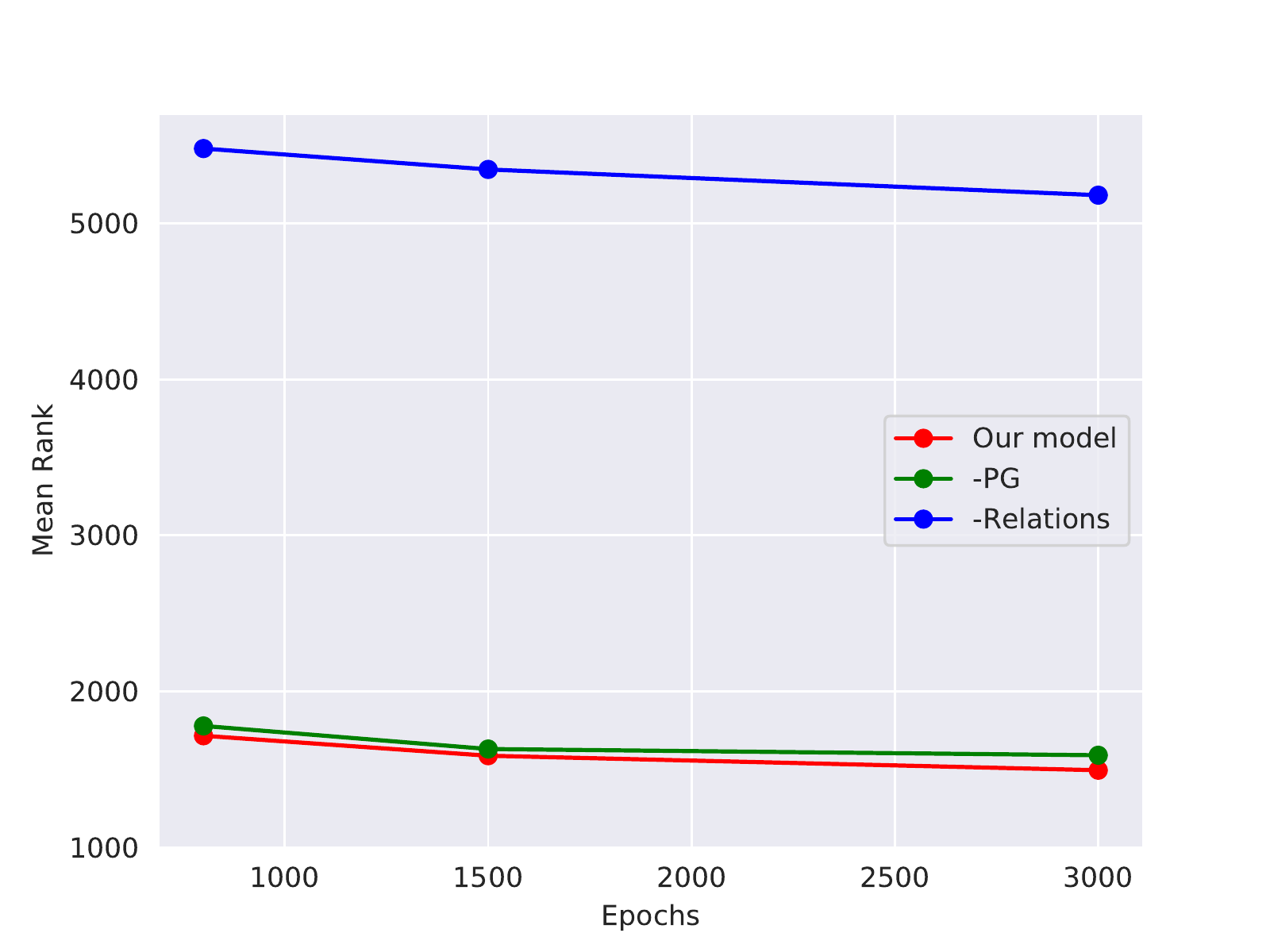}
  \caption{Epochs vs Mean Rank for our model and two ablated models on \emph{NELL-995}. \(-\)PG (green) represents the model after removing \(n\)-hop auxiliary relations or \emph{path generalization}, \(-\)Relations (blue) represents model without taking relations into account and Our model (red) represents the entire model.}
  \label{fig:abl}
\end{figure}

We carry out an ablation study on our model, where we analyze the behavior of \emph{mean rank} on a test set when we omit \emph{path generalization} (\(-\)PG), i.e., removing \(n\)-hop information, and omit \emph{relation Information} (\(-\)Relations) from our model. Figure \ref{fig:abl} shows that our model performs better than the two ablated models and we see a significant drop in the results when using ablated models on \emph{NELL-995}. Removing the relations from the proposed model has a huge impact on the results which suggests that the relation embeddings play a pivotal role in relation prediction.  

\begin{table}[t!]
\centering
\begin{tabular}{l|cc}
\textbf{Dataset}   &  \textbf{PageRank}  & \textbf{Relative Increase} \\ 
\hline
\emph{NELL-995}  & 1.32 & 0.025  \\
\emph{WN18RR}    & 2.44 & -0.01   \\ 
\emph{FB15k-237} & 6.87 & 0.237  \\ 
\emph{UMLS}      & 740    & 0.247   \\
\emph{Kinship}   & 961    & 0.388   \\
\hline
\end{tabular}
\caption{Mean PageRank \(\times 10^{-5}\) vs relative increase in MRR wrt. DistMult. 
  }\label{tb:ablation}
\end{table}

\section{Conclusion and Future Work}\label{conclusion}
\label{ssec:conclusion}
In this paper, we propose a novel approach for relation prediction. Our approach improves over the state-of-the-art models by significant margins. 
Our proposed model learns new graph attention-based embeddings that specifically cater to relation prediction on KGs. Additionally, we generalize and extend graph attention mechanisms to capture both entity and relation features in a multi-hop neighborhood of a given entity. Our detailed and exhaustive empirical analysis gives more insight into our method's superiority for relation prediction on KGs. The proposed model can be extended to learn embeddings for various tasks using KGs such as dialogue generation~\cite{Hehe2017, keizer2017}, and question answering~\cite{zhang2016question,diefenbach2018wdaqua}.

In the future, we intend to extend our method to better perform on hierarchical graphs and capture higher-order relations between entities (like motifs) in our graph attention model.

\section*{Acknowledgments}

We gratefully acknowledge the support of NVIDIA Corporation with the donation of the Titan Xp GPU used for this research. \\


\begin{thebibliography}{29}
\expandafter\ifx\csname natexlab\endcsname\relax\def\natexlab#1{#1}\fi

\bibitem[{Berant~et~al.(2015)Berant,~Alon,~Dagan,~and
  Goldberger}]{berant-etal-2015-efficient}
Jonathan Berant, Noga Alon, Ido Dagan, and Jacob Goldberger. 2015.
\newblock \href {https://doi.org/10.1162/COLI_a_00220} {Efficient global
  learning of entailment graphs}.
\newblock \emph{Computational Linguistics}, 41(2):221--263.

\bibitem[{Berant~et~al.(2013)Berant,~Chou,~Frostig,~and
  Liang}]{berant2013semantic}
Jonathan Berant, Andrew Chou, Roy Frostig, and Percy Liang. 2013.
\newblock Semantic parsing on freebase from question-answer pairs.
\newblock In \emph{Proceedings of the 2013 Conference on Empirical Methods in
  Natural Language Processing (EMNLP), 2013}, pages 1533--1544.

\bibitem[{Berant~et~al.(2010)Berant,~Dagan,~and
  Goldberger}]{berant-etal-2010-global}
Jonathan Berant, Ido Dagan, and Jacob Goldberger. 2010.
\newblock \href {https://www.aclweb.org/anthology/P10-1124} {Global learning of
  focused entailment graphs}.
\newblock In \emph{Proceedings of the 48th Annual Meeting of the Association
  for Computational Linguistics}, pages 1220--1229, Uppsala, Sweden.
  Association for Computational Linguistics.

\bibitem[{Berant~and~Liang(2014)}]{berant2014}
Jonathan Berant and Percy Liang. 2014.
\newblock Semantic parsing via paraphrasing.
\newblock In \emph{Proceedings of the 52nd Annual Meeting of the Association
  for Computational Linguistics (ACL), 2014}.

\bibitem[{Bordes~et~al.(2013)Bordes,~Usunier,~Garcia-Duran,~Weston,~and
  Yakhnenko}]{NIPS2013_5071}
Antoine Bordes, Nicolas Usunier, Alberto Garcia-Duran, Jason Weston, and Oksana
  Yakhnenko. 2013.
\newblock Translating embeddings for modeling multi-relational data.
\newblock In C.~J.~C. Burges, L.~Bottou, M.~Welling, Z.~Ghahramani, and K.~Q.
  Weinberger, editors, \emph{Advances in Neural Information Processing Systems
  (NIPS), 2013}, pages 2787--2795.

\bibitem[{Dettmers~et~al.(2018)Dettmers,~Minervini,~Stenetorp,~and
  Riedel}]{dettmers2018convolutional}
Tim Dettmers, Pasquale Minervini, Pontus Stenetorp, and Sebastian Riedel. 2018.
\newblock Convolutional 2d knowledge graph embeddings.
\newblock In \emph{Thirty-Second AAAI Conference on Artificial Intelligence
  (AAAI), 2018}.

\bibitem[{Diefenbach~et~al.(2018)Diefenbach,~Singh,~and
  Maret}]{diefenbach2018wdaqua}
Dennis Diefenbach, Kamal Singh, and Pierre Maret. 2018.
\newblock Wdaqua-core1: a question answering service for rdf knowledge bases.
\newblock In \emph{Companion of the The Web Conference 2018 on The Web
  Conference (WWW), 2018}, pages 1087--1091. International World Wide Web
  Conferences Steering Committee.

\bibitem[{He~et~al.(2017)He,~Balakrishnan,~Eric,~and~Liang}]{Hehe2017}
He~He, Anusha Balakrishnan, Mihail Eric, and Percy Liang. 2017.
\newblock Learning symmetric collaborative dialogue agents with dynamic
  knowledge graph embeddings.
\newblock In \emph{Proceedings of the 55th Annual Meeting of the Association
  for Computational Linguistics (ACL), 2017}.

\bibitem[{Keizer~et~al.(2017)Keizer,~Guhe,~Cuayahuitl,~Efstathiou,~Engelbrecht,
  Dobre, Lascarides, and Lemon}]{keizer2017}
Simon Keizer, Markus Guhe, Heriberto Cuayahuitl, Ioannis Efstathiou,
  Klaus-Peter Engelbrecht, Mihai Dobre, Alex Lascarides, and Oliver Lemon.
  2017.
\newblock Evaluating persuasion strategies and deep reinforcement learning
  methods for negotiation dialogue agents.
\newblock In \emph{Proceedings of the 15th Conference of the European Chapter
  of the Association for Computational Linguistics (ACL), 2017}.

\bibitem[{Kipf~and~Welling(2017)}]{kipf2017semi}
Thomas~N. Kipf and Max Welling. 2017.
\newblock Semi-supervised classification with graph convolutional networks.
\newblock In \emph{International Conference on Learning Representations (ICLR),
  2017}.

\bibitem[{Kok~and~Domingos(2007)}]{Kok:2007:SPI:1273496.1273551}
Stanley Kok and Pedro Domingos. 2007.
\newblock Statistical predicate invention.
\newblock In \emph{Proceedings of the 24th International Conference on Machine
  Learning}, (ICML), 2007.

\bibitem[{KOTLERMAN~et~al.(2015)KOTLERMAN,~DAGAN,~MAGNINI,~and
  BENTIVOGLI}]{kotlerman_dagan_magnini_bentivogli_2015}
LILI KOTLERMAN, IDO DAGAN, BERNARDO MAGNINI, and LUISA BENTIVOGLI. 2015.
\newblock \href {https://doi.org/10.1017/S1351324915000108} {Textual entailment
  graphs}.
\newblock \emph{Natural Language Engineering}, 21(5):699–724.

\bibitem[{Lin~et~al.(2018)Lin,~Socher,~and~Xiong}]{LinRX2018}
Xi~Victoria Lin, Richard Socher, and Caiming Xiong. 2018.
\newblock Multi-hop knowledge graph reasoning with reward shaping.
\newblock In \emph{Proceedings of the 2018 Conference on Empirical Methods in
  Natural Language Processing (EMNLP), 2018}.

\bibitem[{Lin~et~al.(2015)Lin,~Liu,~Luan,~Sun,~Rao,~and~Liu}]{Lin2015}
Yankai Lin, Zhiyuan Liu, Huanbo Luan, Maosong Sun, Siwei Rao, and Song Liu.
  2015.
\newblock Modeling relation paths for representation learning of knowledge
  bases.
\newblock In \emph{Proceedings of the 2015 Conference on Empirical Methods in
  Natural Language Processing (EMNLP), 2015}.

\bibitem[{Nguyen~et~al.(2018)Nguyen,~Nguyen,~Nguyen,~and
  Phung}]{nguyen2018novel}
Dai~Quoc Nguyen, Tu~Dinh Nguyen, Dat~Quoc Nguyen, and Dinh Phung. 2018.
\newblock A novel embedding model for knowledge base completion based on
  convolutional neural network.
\newblock In \emph{Proceedings of the 2018 Conference of the North American
  Chapter of the Association for Computational Linguistics: Human Language
  Technologies (NAACL), 2018}, volume~2, pages 327--333.

\bibitem[{Nickel~et~al.(2016)Nickel,~Rosasco,~Poggio
  et~al.}]{nickel2016holographic}
Maximilian Nickel, Lorenzo Rosasco, Tomaso~A Poggio, et~al. 2016.
\newblock Holographic embeddings of knowledge graphs.
\newblock In \emph{(AAAI), 2016}, volume~2, pages 3--2.

\bibitem[{Nickel~et~al.(2011)Nickel,~Tresp,~and~Kriegel}]{Nickel2011}
Maximilian Nickel, Volker Tresp, and Hans-Peter Kriegel. 2011.
\newblock A three-way model for collective learning on multi-relational data.
\newblock In \emph{Proceedings of the 28th International Conference on
  International Conference on Machine Learning}, (ICML), 2011.

\bibitem[{Schlichtkrull~et~al.(2018)Schlichtkrull,~Kipf,~Bloem,~van~den~Berg,
  Titov, and Welling}]{schlichtkrull2018modeling}
Michael Schlichtkrull, Thomas~N Kipf, Peter Bloem, Rianne van~den Berg, Ivan
  Titov, and Max Welling. 2018.
\newblock Modeling relational data with graph convolutional networks.
\newblock In \emph{European Semantic Web Conference (ESWC), 2018}, pages
  593--607.

\bibitem[{Socher~et~al.(2013{\natexlab{a}})Socher,~Chen,~Manning,~and
  Ng}]{Socher2013}
Richard Socher, Danqi Chen, Christopher~D Manning, and Andrew Ng.
  2013{\natexlab{a}}.
\newblock Reasoning with neural tensor networks for knowledge base completion.
\newblock In C.~J.~C. Burges, L.~Bottou, M.~Welling, Z.~Ghahramani, and K.~Q.
  Weinberger, editors, \emph{Advances in Neural Information Processing Systems
  26 (NIPS), 2013}, pages 926--934.

\bibitem[{Socher~et~al.(2013{\natexlab{b}})Socher,~Chen,~Manning,~and
  Ng}]{socher2013reasoning}
Richard Socher, Danqi Chen, Christopher~D Manning, and Andrew Ng.
  2013{\natexlab{b}}.
\newblock Reasoning with neural tensor networks for knowledge base completion.
\newblock In \emph{Advances in neural information processing systems (NIPS),
  2013}, pages 926--934.

\bibitem[{Toutanova~et~al.(2015)Toutanova,~Chen,~Pantel,~Poon,~Choudhury,~and
  Gamon}]{toutanova2015representing}
Kristina Toutanova, Danqi Chen, Patrick Pantel, Hoifung Poon, Pallavi
  Choudhury, and Michael Gamon. 2015.
\newblock Representing text for joint embedding of text and knowledge bases.
\newblock In \emph{Proceedings of the 2015 Conference on Empirical Methods in
  Natural Language Processing (EMNLP), 2015}, pages 1499--1509.

\bibitem[{Trouillon~et~al.(2016)Trouillon,~Welbl,~Riedel,~Gaussier,~and
  Bouchard}]{trouillon2016complex}
Th{\'e}o Trouillon, Johannes Welbl, Sebastian Riedel, {\'E}ric Gaussier, and
  Guillaume Bouchard. 2016.
\newblock Complex embeddings for simple link prediction.
\newblock In \emph{International Conference on Machine Learning (ICML), 2016},
  pages 2071--2080.

\bibitem[{Valverde-Rebaza~and~de~Andrade~Lopes(2012)}]{Valverde_2012}
Jorge~Carlos Valverde-Rebaza and Alneu de~Andrade~Lopes. 2012.
\newblock Link prediction in complex networks based on cluster information.
\newblock In \emph{Proceedings of the 21st Brazilian Conference on Advances in
  Artificial Intelligence}, (SBIA), 2012, pages 92--101.

\bibitem[{Vaswani~et~al.(2017)Vaswani,~Shazeer,~Parmar,~Uszkoreit,~Jones,
  Gomez, Kaiser, and Polosukhin}]{NIPS2017_7181}
Ashish Vaswani, Noam Shazeer, Niki Parmar, Jakob Uszkoreit, Llion Jones,
  Aidan~N Gomez, \L~ukasz Kaiser, and Illia Polosukhin. 2017.
\newblock Attention is all you need.
\newblock In I.~Guyon, U.~V. Luxburg, S.~Bengio, H.~Wallach, R.~Fergus,
  S.~Vishwanathan, and R.~Garnett, editors, \emph{Advances in Neural
  Information Processing Systems (NIPS), 2017}, pages 5998--6008.

\bibitem[{Veli{\v{c}}kovi{\'{c}}~et~al.(2018)Veli{\v{c}}kovi{\'{c}},~Cucurull,
  Casanova, Romero, Li{\`{o}}, and Bengio}]{velickovic2018graph}
Petar Veli{\v{c}}kovi{\'{c}}, Guillem Cucurull, Arantxa Casanova, Adriana
  Romero, Pietro Li{\`{o}}, and Yoshua Bengio. 2018.
\newblock {Graph Attention Networks}.
\newblock \emph{International Conference on Learning Representations (ICLR),
  2018}.

\bibitem[{West~et~al.(2014)West,~Gabrilovich,~Murphy,~Sun,~Gupta,~and
  Lin}]{West_2014}
Robert West, Evgeniy Gabrilovich, Kevin Murphy, Shaohua Sun, Rahul Gupta, and
  Dekang Lin. 2014.
\newblock Knowledge base completion via search-based question answering.
\newblock In \emph{Proceedings of the 23rd International Conference on World
  Wide Web}, (WWW), 2014, pages 515--526.

\bibitem[{Xiong~et~al.(2017)Xiong,~Hoang,~and~Wang}]{xiong2017}
Wenhan Xiong, Thien Hoang, and William~Yang Wang. 2017.
\newblock Deeppath: A reinforcement learning method for knowledge graph
  reasoning.
\newblock In \emph{Proceedings of the 2017 Conference on Empirical Methods in
  Natural Language Processing (EMNLP), 2017}.

\bibitem[{Yang~et~al.(2015)Yang,~Yih,~He,~Gao,~and~Deng}]{yang2014}
Bishan Yang, Wen-tau Yih, Xiaodong He, Jianfeng Gao, and Li~Deng. 2015.
\newblock {Embedding Entities and Relations for Learning and Inference in
  Knowledge Bases}.
\newblock \emph{International Conference on Learning Representations (ICLR),
  2015}.

\bibitem[{Zhang~et~al.(2016)Zhang,~Liu,~He,~Ji,~Liu,~Wu,~and
  Zhao}]{zhang2016question}
Yuanzhe Zhang, Kang Liu, Shizhu He, Guoliang Ji, Zhanyi Liu, Hua Wu, and Jun
  Zhao. 2016.
\newblock Question answering over knowledge base with neural attention
  combining global knowledge information.
\newblock \emph{arXiv preprint arXiv:1606.00979, 2016}.

\end{thebibliography}

\end{document}